\documentclass[letterpaper, 10 pt, conference]{ieeeconf}

\IEEEoverridecommandlockouts                              %

\usepackage{amsmath,amsfonts}
\usepackage{array}
\usepackage[caption=false,font=footnotesize,labelfont=sf,textfont=sf]{subfig}
\usepackage{textcomp}
\usepackage{stfloats}
\usepackage{url}
\usepackage{verbatim}
\usepackage{graphicx}
\def\BibTeX{{\rm B\kern-.05em{\sc i\kern-.025em b}\kern-.08em
    T\kern-.1667em\lower.7ex\hbox{E}\kern-.125emX}}
\usepackage{balance}
\usepackage{hyperref}

\usepackage{xcolor}
\usepackage{amsmath,amsfonts,bm,bbm,mathtools,amssymb}
\usepackage{physics}
\usepackage{cleveref}
\crefname{algocf}{algorithm}{algorithms}
\Crefname{algocf}{Algorithm}{Algorithms}
\usepackage{csquotes}
\usepackage{anyfontsize}  %
\usepackage{tikz}

\usepackage{multirow}
\usepackage{booktabs} %

\usepackage{times}

\usepackage{newtxtext}

\usepackage{wrapfig}

\usepackage{adjustbox}

\usepackage[font=small]{caption}

\usepackage[caption=false,font=footnotesize]{subfig}
\usepackage{cite}

\newcommand{\citet}{\cite}

\def\eqref#1{equation~\ref{#1}}

\usepackage[ruled,vlined,linesnumbered]{algorithm2e}

\usepackage{setspace} 

\SetCommentSty{mycommfont}
\SetKwComment{Comment}{$\triangleright$\ }{}
\let\oldnl\nl%

\newcommand{\nonl}{\renewcommand{\nl}{\let\nl\oldnl}}%
\SetAlFnt{\small}           %
\SetAlCapNameFnt{\small}    %
\SetAlCapFnt{\small}        %

\def\1{\bm{1}}

\DeclareMathOperator*{\argmax}{arg\,max}

\def\vmu{{\bm{\mu}}}

\def\vtheta{{\bm{\theta}}}
\def\vepsilon{{\bm{\epsilon}}}

\def\va{{\bm{a}}}
\def\vb{{\bm{b}}}
\def\vc{{\bm{c}}}

\def\ve{{\bm{e}}}

\def\vg{{\bm{g}}}

\def\vp{{\bm{p}}}
\def\vq{{\bm{q}}}

\def\vt{{\bm{t}}}

\def\vv{{\bm{v}}}

\def\vx{{\bm{x}}}

\def\vz{{\bm{z}}}

\def\mG{{\bm{G}}}
\def\mH{{\bm{H}}}
\def\mI{{\bm{I}}}

\def\mR{{\bm{R}}}

\def\mSigma{{\bm{\Sigma}}}

\DeclareMathAlphabet{\mathsfit}{\encodingdefault}{\sfdefault}{m}{sl}
\SetMathAlphabet{\mathsfit}{bold}{\encodingdefault}{\sfdefault}{bx}{n}

\newcommand{\pdata}{p_{\rm{data}}}

\newcommand{\E}[2]{\mathbb{E}_{#1}\left[#2\right]}

\newcommand{\R}{\mathbb{R}}

\newcommand{\tran}{^\top}

\newcommand{\SEthree}{\textsc{SE(3)}}

\newcommand{\SOthree}{\textsc{SO(3)}}

\newcommand{\SOthreeRthree}{\textsc{SO(3)} \times \R^3}

\newcommand{\q}{\vq}

\newcommand{\HBinA}[2]{{}^{#1}\mH^{#2}}

\newcommand{\Gaussian}[1]{\mathcal{N}\left(#1\right)}

\newcommand{\alphacumprod}{\bar{\alpha}}
\newcommand{\betaposterior}{\tilde{\beta}}

\newcommand{\relu}{\text{ReLU}}

\newcommand{\objective}{\mathcal{O}}
\newcommand{\context}{\mathcal{C}}

\newcommand{\IGSOthree}{\mathcal{IG}_{\SOthree}}

\newcommand{\IGSOthreeRthree}{\mathcal{IG}_{\SOthree \times \R^3}}

\DeclareMathOperator{\Exp}{Exp}
\DeclareMathOperator{\Log}{Log}

\newcommand{\grasp}{\mathcal{G}}

\graphicspath{
{./content/introduction/figures/}
{./content/experiments/figures/}
{./content/method/figures/}
}

\begin{document}
\bstctlcite{IEEEexample:BSTcontrol}  %

\title{
\textsc{Grasp Diffusion Network}:
Learning Grasp Generators from \\
Partial Point Clouds with Diffusion Models in $\SOthree \times \R^3$
}

\author{
Joao Carvalho$^{1}$,
An T. Le$^{1}$,
Philipp Jahr$^{1}$,
Qiao Sun$^{1}$,
Julen Urain$^{3}$,
Dorothea Koert$^{1,2}$,
and
Jan Peters$^{1,3,4}$%
\thanks{
Corresponding author: Jo\~{a}o Carvalho, \href{joao@robot-learning.de}{joao@robot-learning.de}
}%
\thanks{
This work was funded by the German Federal Ministry of Education and Research projects IKIDA (01IS20045) and Software Campus project ROBOSTRUCT (01S23067), and by the German Research Foundation project METRIC4IMITATION (PE 2315/11-1).
$^{1}$Intelligent Autonomous Systems Lab, Computer Science Department, Technical University of Darmstadt, Germany;
$^{2}$Centre for Cognitive Science, Technical University of Darmstadt, Germany;
$^{3}$German Research Center for AI (DFKI), Research Department: SAIROL, Darmstadt, Germany; 
$^{4}$Hessian.AI, Darmstadt, Germany
}%
}

\maketitle

\begin{abstract}
Grasping objects successfully from a single-view camera is crucial in many robot manipulation tasks.
An approach to solve this problem is to leverage simulation to create large datasets of pairs of objects and grasp poses, and then learn a conditional generative model that can be prompted quickly during deployment.
However, the grasp pose data is highly multimodal since there are several ways to grasp an object.
Hence, in this work, we learn a grasp generative model with diffusion models to sample candidate grasp poses given a partial point cloud of an object.
A novel aspect of our method is to consider diffusion in the manifold space of rotations and to propose a collision-avoidance cost guidance to improve the grasp success rate during inference.
To accelerate grasp sampling we use recent techniques from the diffusion literature to achieve faster inference times.
We show in simulation and real-world experiments that our approach can grasp several objects from raw depth images with $90\%$ success rate and benchmark it against several baselines.
\href{https://sites.google.com/view/graspdiffusionnetwork}{https://sites.google.com/view/graspdiffusionnetwork}
\end{abstract}

\section{Introduction}
\label{sec:introduction}

With the increasing amounts of available simulated and real-world data~\cite{DBLP:conf/icra/ONeillRMGPLPGMJ24}, there is a strong demand for developing deep learning models that can encode and represent high-dimensional and highly multimodal data.
In recent years, diffusion models have been shown to have these desirable properties. 
They are prolific in many robotic manipulation fields~\cite{urain2024deepgenerativemodelsrobotics}, for imitation learning~\cite{DBLP:conf/rss/ChiFDXCBS23,DBLP:conf/rss/ReussLJL23}, grasping~\cite{urain2022se3diffusion} and motion planning~\cite{DBLP:conf/iros/Carvalho0BK023}.

In this work, given a \textit{partial view} of an object from a depth camera (a point cloud), we consider the problem of learning a distribution of \textit{good} grasps.
A \textit{good} grasp means the gripper can grasp and lift an object successfully without falling from its fingers.
Full point clouds are often impractical to obtain in the real world, especially for grasping scenarios (e.g., if an object is placed on top of a table, we cannot obtain the point cloud of its bottom).
Moreover, multiple cameras or views are needed to obtain a complete point cloud.
Both are undesirable for robots that need to operate fast and are cheap to deploy (i.e., using only one camera).

\begin{figure}[!t]
  \centering
  \includegraphics[width=0.75\linewidth]{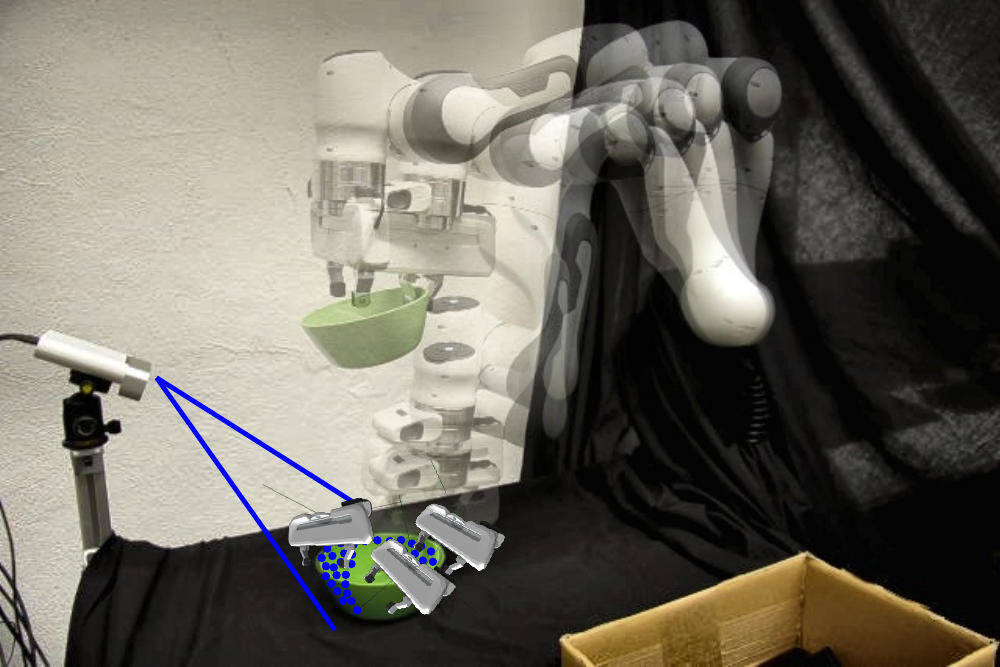}
  \caption[Real world experiment setup for grasping experiments.]{
    Setup for real-world experiments and overlay execution of a successful grasp.
    Given an object partial point cloud (blue dots), GDN generates multimodal grasps and executes one successfully.
  }
  \label{fig:grasping_real_world}
  \vspace{-0.75cm}
\end{figure}

We assume having a parallel gripper, where a homogeneous transformation in the space of rotations and translations represents a grasp pose.
These grippers are commonly available and cheaper than more dexterous robot hands, and they are sufficient for many pick-and-place tasks.
The grasp pose data is often not unimodal since several grasp poses can successfully grasp an object.
Due to several desirable properties, diffusion models are a good candidate for modeling grasp distributions.
First, they have been shown to model multimodal distributions quite well,
and avoid mode collapse commonly present in other generative models, such as
VAEs and GANs~\cite{kingma2013vae,goodfellow2014gan}.
Second, they are empirically more stable to train than other generative models~\cite{DBLP:conf/iclr/Bayat23}.
Third, they can encode large amounts of data, as the simulated grasp dataset we use~\cite{DBLP:conf/icra/EppnerMF21}.
Finally, 
diffusion models can be used as priors to sample from posterior distributions that combine merging the prior with a differentiable likelihood - commonly known as classifier-guidance~\cite{Dhariwal2021diffusionbeatsgans}.

Denoising Diffusion Probabilistic Models (DDPM)~\cite{ho2020denoisingdiffusion} were designed for data belonging to an Euclidean space, and thus, a Gaussian distribution in $\R^n$ is used. 
However, grasp poses belong to the space of homogeneous transformations $\SEthree$.
To adapt diffusion to this space, we decouple the Lie group $\SEthree$ into $\SOthree \times \R^3$ and learn a diffusion model in the Lie algebra.
As previously shown~\cite{leach2022denoising,DBLP:conf/aaai/JagvaralLM24}, properly modeling the diffusion of rotations has several benefits.

Our contributions include:
(1) We present Grasp Diffusion Network (GDN) -- a new grasp generative model conditioned on partial point clouds that uses diffusion in the $\SOthreeRthree$ manifold.
(2) To increase the success rate, we propose sampling from a posterior distribution that combines the learned grasp prior with a collision-avoidance cost.
(3) We perform experiments both in simulation and real-world scenarios
using partial point clouds obtained
with an external RGB-D camera, and show that GDN achieves higher success rates and closeness to the data distribution than the baselines.

\section{Related Work}
\label{sec:related_work}

\textbf{Grasp Generative Methods.}
Several works model the distribution of grasps given raw point cloud inputs of objects or scenes.
For complete surveys on learning and deep learning models for grasping, we refer the readers to~\cite{DBLP:journals/trob/NewburyGCMELBMAKFC23,DBLP:journals/arcras/Platt23}.
A seminal work in learning grasp generative models is $6$-dof GraspNet~\cite{DBLP:conf/iccv/MousavianEF19}, which uses a conditional variational autoencoder (CVAE) to learn a grasp distribution given partial point clouds from objects observations.
In that work, the network's encoder and decoder parts include point cloud encoder modules.
Our approach only has one point cloud encoder, which works as a conditioning variable.
VAEs are trained to compress and reconstruct the grasps and are known for showing mode-collapse and not fully capturing the data distribution.
Moreover, the authors learn both a grasp generator on \textit{good} grasps and a grasp evaluator, implemented as a classifier to discern between \textit{good} and \textit{bad} grasps.
However, ideally, the generative model should already model the \textit{good} grasps distribution well.
Contact-GraspNet~\cite{DBLP:conf/icra/SundermeyerMTF21} moves grasp generation from a single object to a full scene point cloud to produce grasp poses that avoid collisions with the scene.
\citet{DBLP:conf/icra/MuraliMEPF20} extends $6$-dof GraspNet to full scenes represented with partial point clouds and generates grasp poses by segmenting target objects and ranking solutions by their collisions with the scene.
Instead of predicting grasp distributions for the whole object, \citet{DBLP:conf/iros/LundellVLMFK23} learns a distribution using a CVAE that is conditioned on the object point cloud and an area of interest on the object surface.

\textbf{Diffusion-based Grasp Generative Methods.}
As the grasp distribution is highly multimodal, in recent works, score-based and diffusion models have been proposed~\cite{urain2022se3diffusion,song2024implicit,DBLP:journals/corr/abs-2312-11243}.
$\SEthree$-DiffusionFields~\cite{urain2022se3diffusion} learns a grasp distribution of parallel grippers with a score-based model. 
The authors learn an energy-based model (EBM) via denoising score matching,
by matching the energy function's gradient with the log-probability derivative in the tangent space of $\SEthree$, and use Langevin dynamics
to sample from the learned model.
Computing the log-probability gradient involves backpropagating through the entire EBM network.
Instead, we model the generative model using DDPM, and our network directly computes the gradient as the denoising function output, which is common practice in diffusion models.
This has several advantages, such as saving computation time (during training and inference) and training stability.
GraspLDM~\cite{DBLP:journals/corr/abs-2312-11243} learns a grasp generative model using latent diffusion in Euclidean space and uses language to condition where to grasp the object (e.g., top or bottom).
However, a CVAE must first be trained to learn the latent space representation.
Similar to ours, their CVAE baseline results show already satisfactory success rates.
Several works have also combined grasp diffusion with language~\cite{DBLP:journals/corr/abs-2407-13842}.
To deal with point cloud input equivariance, such that rotating an object should also rotate the associated grasp poses, \citet{lim2024leveraging}~proposed using a VNN equivariant point cloud encoder~\cite{deng2021vnn} and flow-matching~\cite{DBLP:conf/iclr/LipmanCBNL23}.0
However, this work shows results for full point clouds, which are more impractical in the real world.

\section{Grasp Diffusion Network}
\label{sec:grasp_diffusion_network}

In this work, we consider the problem of generating a distribution of \textit{good} grasps $\grasp$ conditioned on a partial point cloud observation $p(\grasp | \context)$, where $\context$ is a context variable representing the point cloud.
Since this distribution can be highly multimodal, 
we model it with diffusion models. 
We consider the widely available parallel grippers and represent a grasp pose ${\mG = (\vt, \mR)}$ as an element of the $\SEthree$ Lie group of homogeneous transformations, where $\vt \in \R^3$ is a translation vector and $\mR \in \SOthree$ a rotation.
To convert elements between Lie groups and the Lie algebras vector representations, we use the logarithmic ($\Log$) and the exponential ($\Exp$) maps.
See~\cite{sola2018micro} for details.

Diffusion models are a class of implicit generative models~\cite{ho2020denoisingdiffusion},
which perturb the original data distribution $\pdata(\mG_{0} | \context)$ via a noise process and learn to reconstruct it with denoising.
As DDPM models were formally introduced for the Euclidean space, for the translation part, we use DDPM in $\R^3$, while for the rotation part, we define a diffusion model in $\SOthree$ as in~\cite{leach2022denoising}.
We drop the conditioning variable in the following sections for a clearer notion.

\subsection{Diffusion for Translations in \texorpdfstring{$\R^3$}{R3}}

Let $\vt_0 \in \R^n$ be a translation sampled from the data distribution and transformed into white Gaussian noise by a Markovian forward diffusion process 
in $N$ steps
${
q(\vt_{i} | \vt_{i-1}, i) = \Gaussian{\vt_{i}; \sqrt{1-\beta_i} \vt_{i-1}, \beta_i \mI}
}$, 
where $i$ is the diffusion time step, $\beta_i$ is the noise scale at time step $i$, and $q(\vt_N) \approx \Gaussian{ \vt_N; \bm{0}, \mI}$.
We use the cosine noise schedule for $\beta$~\cite{ho2020denoisingdiffusion,nichol2021improvedddpm}.
The denoising process transforms noise back to the data distribution such that $p(\vt_0) \approx q(\vt_0)$, utilizing a reverse Markov process
$
p(\vt_{0:N}) = p(\vt_N) \prod_{i=1}^{N} p(\vt_{i-1} | \vt_i, i)
$
Sampling from $p(\vt_0)$ is done by starting from a sample from an isotropic Gaussian in $\R^3$ and sequentially sampling from the posterior distribution $p(\vt_{i-1} | \vt_i, i)$ at each denoising step $i$.
For $N \to \infty$ and $\beta_i \to 0$, this posterior converges to a Gaussian distribution~\cite{Sohl-Dickstein2015diffusion}.
Hence, during training, the goal is to learn to approximate a Gaussian posterior
\begin{align}
    \label{eq:ddpm_denoising_posterior}
    p_{\vtheta}(\vt_{i-1} | \vt_i, i) = \Gaussian{\vt_{i-1}; \vmu_i = \vmu_{\vtheta}(\vt_{i}, i), \mSigma_i}.
\end{align}
For simplicity~\cite{ho2020denoisingdiffusion}, only the mean of the inverse process is learned, and the covariance is set to
${\mSigma_i = \sigma_i^2 \mI = \betaposterior_i \mI}$,
with ${\betaposterior_i = (1-\alphacumprod_{i-1})/(1-\alphacumprod_i) \beta_i}$
and
${\alpha_i=1-\beta_i}$ and ${\alphacumprod_i = \prod_{k=1}^i \alpha_k}$.
Instead of learning the posterior mean directly, \citet{ho2020denoisingdiffusion}~proposed to learn the noise vector $\vepsilon_{\vtheta}$, since
\begin{align}
    \label{eq:diffusion_posterior_mean}
    \vmu_{\vtheta}(\vt_i, i) = \frac{1}{\sqrt{\alpha_i}} \left( \vt_i - \frac{1 - \alpha_i}{\sqrt{1-\alphacumprod_i}} \vepsilon_{\vtheta}(\vt_i, i) \right).
\end{align}
The parameters are learned by maximizing an evidence lower bound (ELBO) on the expected data log-likelihood
$\argmax_{\vtheta} \E{\vt_0 \sim \pdata}{\log p_{\theta}(\vt_0)}$,
which after simplifying~\cite{ho2020denoisingdiffusion}, the parameters $\vtheta$ are learned by minimizing
\begin{align}
    \label{eq:diffusion_loss_translation}
    &\mathcal{L}_{\vt}(\vtheta) = \E{i, \vepsilon, \vt_0 }{ \| \vepsilon - \vepsilon_{\vtheta}(\vt_i, i) \|_2^2 } \\
    &i \sim \mathcal{U}(1, N), \, \vepsilon \sim \Gaussian{\bm{0}, \mI} , \,
    \vt_0 \sim q(\vt_0), \, \vt_i \sim q(\vt_{i}|\vt_0, i) \nonumber
\end{align}
The distribution of the noisy sample at time step $i$ is Gaussian
$q(\vt_{i}|\vt_0, i) = \Gaussian{\vt_i; \sqrt{\alphacumprod_i}\vt_0, (1-\alphacumprod_i) \mI}$,
which allows for efficient training by sampling it without running the forward diffusion process.
The derivation of the simplified loss in \cref{eq:diffusion_loss_translation} from the ELBO involves computing the KL divergence
$D_{\text{KL}} \left( q(\vt_{i-1} | \vt_{i}, \vt_{0}) \|  p_{\vtheta}(\vt_{i-1} | \vt_{i}) \right)$,
which has a closed-form solution if the distribution 
$q(\vt_{i-1} | \vt_{i}, \vt_{0}) = q(\vt_{i} | \vt_{i-1}, \vt_{0}) q(\vt_{i-1} | \vt_{0}) / q(\vt_{i} | \vt_{0})$
is Gaussian,
which it is, because all on the right-hand side are Gaussian~\cite{pml1Book}.
After training,
we obtain a sample from $p_{\vtheta}(\vt_0)$ by
iteratively sampling from
\cref{eq:ddpm_denoising_posterior}
for $N$ denoising steps.

\subsection{Diffusion for Rotations in \texorpdfstring{$\SOthree$}{SO(3)}}

The Euclidean diffusion model does not directly apply to rotations since these are elements of $\SOthree$ and not $\R^n$.
Alternatively, one could build a diffusion model
using Euler-angles or quaternions, but these representations have issues of not representing the diffusion distribution on $\SOthree$ correctly~\cite{leach2022denoising}.
Euler angles suffer from Gimbal lock.
Quaternions reside on the unit norm ($S^3$ sphere), and during optimization, they can deviate from the unit sphere, leading to invalid rotations, and need to be projected back to the space of rotations at every diffusion step.
Hence, following~\cite{leach2022denoising,DBLP:conf/aaai/JagvaralLM24}, we consider the rotation diffusion process on the isotropic Gaussian in ${\SOthree - \IGSOthree}$,
which is defined by a mean rotation $\vmu \in \SOthree$ and a scale $\epsilon \in \R$.
The probability density of a rotation $\mR \in \SOthree$ is
\begin{equation*}
    \IGSOthree(\mR; \vmu, \epsilon) = f_{\epsilon} \left( \arccos\left( 1/2 \, \text{tr}(\vmu\tran \mR) - 1 \right) \right).
\end{equation*}
This distribution can be parametrized in the axis-angle form.
The axis follows a uniform and normalized distribution on the sphere $\mathcal{S}^2$,
and the density of a rotation angle $\omega \in [0, \pi)$ is modeled with $f_{\epsilon}(\omega)$.
Sampling a rotation ${\mR \sim \IGSOthree(\cdot; \vmu, \epsilon)}$ is done by first sampling from $\IGSOthree(\cdot; \mI, \epsilon)$ using the inverse sampling transform method, and then rotating the sample by $\vmu$.
See~\cite{leach2022denoising,DBLP:conf/aaai/JagvaralLM24} for details.

The motivation for using $\IGSOthree$ in diffusion for rotations is twofold.
(1) DDPM can be interpreted as a discretized version of a Variance-Preserving (VP) Stochastic Differential Equation (SDE)~\cite{song2021scorebased}.
In the space of rotations, this SDE's solution approaches the $\IGSOthree$ with identity rotation and unity scale~\cite{DBLP:conf/aaai/JagvaralLM24}.
(2) The $\IGSOthree$ distribution is closed under the convolution operation, contrary to the Bingham and Matrix Fisher distributions~\cite{leach2022denoising}.
This property is essential for efficiently training DDPM models without simulating the whole Markov chain.
It is well known that the sum of two independent Gaussian distributed random variables is also a Gaussian distribution~\cite{pml1Book}.
Similarly, 
the multiplication of two independent $\IGSOthree$ random variables also follows an $\IGSOthree$ distribution.

Following~\cite{leach2022denoising}, it is unclear how to obtain their proposed loss function.
Hence, we establish here the equivalent of \cref{eq:diffusion_loss_translation} for rotations.
Let $\lambda$ be the geodesic interpolation from the identity rotation
${\lambda(\gamma, \mR) = \exp(\gamma \log(\mR))}$.
From the convolution property, we have
\begin{align}
    & \mR_0 \sim \pdata(\mR_0), \quad  \mR \sim \IGSOthree \left( \mR ; \mI, \beta_i  \right) \label{eq:rot_eqs} \\
    & \mR_i = \mR \lambda( \sqrt{1-\beta_i}, \mR_{i-1} ) \nonumber \\
    & q(\mR_{i} | \mR_{i-1}) = \IGSOthree \left( \mR_{i}; \lambda( \sqrt{1-\beta_i}, \mR_{i-1} ), \beta_i  \right)  \nonumber \\
    & q(\mR_{i} | \mR_{0}) = \IGSOthree \left( \mR_{i}; \lambda( \sqrt{\alphacumprod_i}, \mR_{0} ), 1 - \alphacumprod_i  \right). \nonumber
\end{align}
Hence, we can sample $\mR_i$ from $q(\mR_{i}|\mR_0, i)$ as
${
\mR_i = \mR \exp(\sqrt{\alphacumprod_i} \log \mR_0)
}$,
with
${
 \mR \sim \IGSOthree (\mI, \sqrt{1 - \alphacumprod_i})
}$.
Similar to DDPM in Euclidean space, in $\SOthree$ the denoising posterior is approximated by learning the mean rotation of an $\IGSOthree$ distribution
\begin{align}
    \label{eq:diffusion_posterior_so3}
    p_{\theta} (\mR_{i-1}|\mR_{i}, i) = \IGSOthree (\mR_{i-1}; \vmu_{\vtheta}(\mR_{i}, i), \mSigma_i).
\end{align}
The mean $\vmu_{\vtheta}(\mR_{i}, i)$ cannot be directly constructed as in \cref{eq:diffusion_posterior_mean} by simply replacing $\vt_i$ with $\mR_i$, since the latter is an element of $\SOthree$ and not $\R^3$.
Consequently, the loss function in \cref{eq:diffusion_loss_translation} cannot be directly applied to rotations because the KL divergence used in the derivation cannot be computed in closed form.
However, for small values of $\epsilon$ the distribution $\IGSOthree(\mR; \vmu, \epsilon)$ can be approximated locally in the tangent space by a Gaussian distribution
$\Gaussian{\mR^{\text{aa}}; \vmu^{\text{aa}}=\Log(\vmu), \sigma^2 \mI}$,
using the axis-angle parametrization with $\vmu^{\text{aa}} \in \R^3$, and $\epsilon = \sigma^2 / 2$~\cite{DBLP:conf/aaai/JagvaralLM24}.
A ``small"~$\epsilon$ means the density function drops to a value close to zero within a sphere around the mean, meaning the local approximation is valid~\cite{DBLP:conf/cdc/ChirikjianK14}.
Since the DDPM's diffusion process is a variance-preserving method,
we have
$\epsilon_i \equiv  \sqrt{1 - \alphacumprod_i}$,
which drops from $1$ to $0$ with $i = 1 \ldots N$, and locally approximate $\IGSOthree$ with a Gaussian in the tangent space using the axis-angle representation.
Applying this approximation to our distributions, we obtain
\begin{align}
    q(\mR_{i} | \mR_{i-1}) & = \IGSOthree \left( \mR_{i}; \lambda \left( \sqrt{1-\beta_i}, \mR_{i-1} \right), \beta_i  \right) \label{eq:igso3_q_noising}\\
    & \approx \Gaussian{\vv_{i}; \sqrt{1-\beta_i} \Log\left( \mR_{i-1} \right), 2 \beta_i \mI},\nonumber
\end{align}
where $\vv_i \in \R^3$ is the representation of $\mR_i$ in the Lie algebra.
A similar analogy can be derived for $q(\mR_{i} | \mR_{0})$.
Since these distributions are locally approximated with Gaussians, 
we can follow the same derivations of DDPM for the Euclidean space
to compute an approximation of
$q(\mR_{i-1} | \mR_{i}, \mR_{0})$
in the tangent space,
and after simplifying, we obtain
\begin{align}
    & \log \vmu_{\vtheta}(\mR_i, i) =  \frac{1}{\sqrt{\alpha_i}} \left( \Log \mR_i - \frac{1-\alpha_i}{\sqrt{1 - \alphacumprod_i}} \vepsilon_{\vtheta}(\mR_i, i) \right) \nonumber \\
    & \vmu_{\vtheta}(\mR_i, i) = \Exp \left( \log \vmu_{\vtheta}(\mR_i, i) \right) \label{eq:diffusion_posterior_mean_so3}
\end{align}
where $\vepsilon_{\vtheta}(\mR_i, i) \in \R^3$ is a vector represented in the Lie algebra of $\SOthree$.
This vector can be interpreted as an angular velocity.
The posterior mean proposed in~\cite{leach2022denoising},
which is a function of the current rotation $\mR_i$ and the estimated denoised rotation $\hat{\mR}_0$,
can be derived from \cref{eq:diffusion_posterior_mean_so3}.

With the local Gaussian approximations in the tangent space, 
the loss derivation is similar to diffusion in Euclidean space,
and thus the loss function is similar to \cref{eq:diffusion_loss_translation}
\begin{align}
    \label{eq:diffusion_loss_so3}
    & \mathcal{L}_{\mR}(\vtheta) = \E{i, \vepsilon, \mR_0 }{ \| \vepsilon - \vepsilon_{\vtheta}(\mR_i, i) \|_2^2 },
    \\
    &
    i \sim \mathcal{U}(1, N), \,
    \mR_0 \sim \q(\mR_0), \,
    \mR \sim \IGSOthree (\mI, \sqrt{1 - \alphacumprod_i}), \,
    \nonumber \\
    &
    \mR_i = \mR \lambda( \sqrt{\alphacumprod_i}, \mR_{0}), \,
    \epsilon = \Log(\mR)/\sqrt{1-\alphacumprod_i}.
    \nonumber
\end{align}
In our implementation, the input $\mR_i$ to the function $\vepsilon_{\vtheta}$ is a flattened rotation matrix in $\R^9$, and the output is the denoising vector in the tangent space at the identity in $\R^3$.
To sample rotations, we follow the DDPM denoising process 
and iteratively sample from the posterior of \cref{eq:diffusion_posterior_so3}, using the learned mean from \cref{eq:diffusion_posterior_mean_so3}, 
or the one proposed in~\cite{leach2022denoising},
which empirically led to slightly better results.

\begin{figure*}[!t]
    \centering

    \subfloat[Grasp Diffusion Network inference pipeline]{%
        \includegraphics[width=0.62\linewidth]{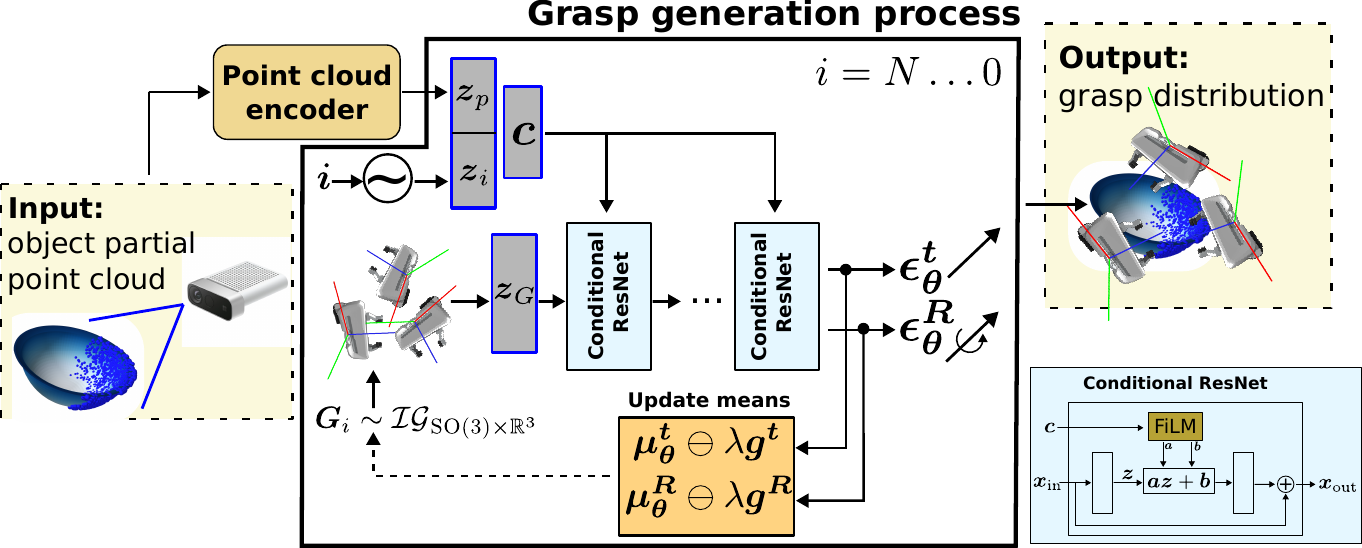}
        \label{fig:gdn_method_overview}
    }
    \hfill
    \subfloat[Samples from GDN with DDPM and DDIM]{%
        \includegraphics[width=0.35\linewidth]{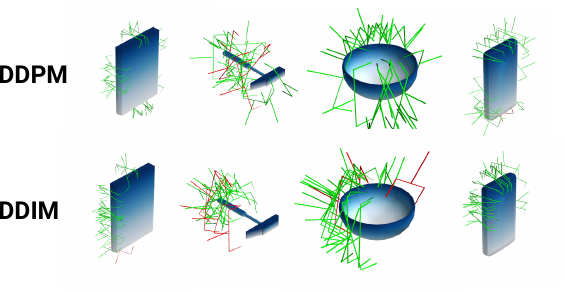}
        \label{fig:grasp_samples_diffusion_ddim}
    }
    \hfill
    \caption[Overview of the Grasp Diffusion Network inference pipeline.]{
        \protect\subref{fig:gdn_method_overview}
        The input is a partial point cloud view of the object to grasp (blue dots),
        and GDN outputs a distribution of gripper poses by denoising in the $\SOthreeRthree$ manifold.
        The denoising network (a conditional ResNet) computes vectors for translation and rotation $\vepsilon^{\vt}_{\vtheta}$ and $\vepsilon^{\mR}_{\vtheta}$ in the Lie algebra.
        These vectors are used to update the means of the posterior distribution, optionally using collision-avoidance cost guidance with the gradients $\vg$.
        \protect\subref{fig:grasp_samples_diffusion_ddim}
        Grasp samples generated with GDN using DDPM and DDIM sampling methods.
        The results align with those from \cref{fig:simulation_results_ddim_cat10}.
        DDIM produces successful grasps (in green) but with less variability.
    }
    \vspace{-0.5cm}
\end{figure*}

\subsection{Diffusion Model for Grasp Poses in \texorpdfstring{$\SEthree$}{SE(3)}}

To learn a diffusion model over grasp poses ${\mG = (\vt, \mR) \in \SEthree}$, we build the posterior as a factorized isotropic Gaussian in $\SOthree$ and $\R^3$
\begin{align*}
    & p_{\vtheta}(\mG_{i-1} | \mG_{i}, i) = \IGSOthreeRthree ( \mG_{i-1}; \vmu_{\vtheta}(\mG_{i}, i), \mSigma_i) \\
    & \; =
    \Gaussian{\vt_{i-1}; \vmu_{\vtheta}^{\vt}(\mG_{i}, i), \mSigma_i}
    \IGSOthree ( \mR_{i-1}; \vmu_{\vtheta}^{\mR}(\mG_{i}, i), \mSigma_i), \nonumber
\end{align*}
where the means for translation and rotation are computed with a joint denoising vector
${
\vepsilon_{\vtheta}(\mG_i, i) = [\vepsilon_{\vtheta}^{\vt}, \vepsilon_{\vtheta}^{\mR}] \in \R^6
}$.

Finally, to build a model conditioned on ${\vc \in \context}$, a partial point cloud of an object, we use a conditional diffusion model
$\pdata(\mG | \vc)$.
The denoising network is extended with an additional input $\vepsilon_{\vtheta}(\mG_i, i, \vc)$, and the expectation includes the context distribution 
${\vc \sim p(\vc)}$, ${\mG_0 \sim q(\mG_0 | \vc)}$.
The parameters $\vtheta$ are learned by optimizing
\begin{align}
    \label{eq:diffusion_loss_se3}
    \mathcal{L}(\vtheta) = \E{i, \vepsilon, \vc, \mG_0 }{ \| \vepsilon - \vepsilon_{\vtheta}(\mG_i, i, \vc) \|_2^2 },
\end{align}
where the noisy translation and rotation inputs are constructed as in \cref{eq:diffusion_loss_translation,eq:diffusion_loss_so3}.
For faster inference, we also use Denoising Diffusion Implicit Models (DDIM)~\cite{DBLP:conf/iclr/SongME21}, which is particularly beneficial in our case because we need to sample from the $\IGSOthree$ distribution at every step.

\subsection{Architecture and Implementation Details}
\label{sec:gdn_architecture_and_implementation_details}

An overview of our method - Grasp Diffusion Network (GDN) - is presented in \cref{fig:gdn_method_overview}.
Given a partial point cloud with $n$ points $\vp \in \R^{n \times 3}$, 
to compute an embedding $\vz_{p} \in \R^{d_p}$ we use the PointNet++ architecture~\cite{DBLP:conf/nips/QiYSG17} with a similar structure as in~\cite{DBLP:conf/iccv/MousavianEF19}.
The diffusion time index $i$ is embedded with a sinusoidal positional embedding $\vz_i \in \R^{d_i}$ and concatenated with the point cloud embedding, providing a context variable 
${\vc = [\vz_p, \vz_i] \in \R^{d_p + d_i}}$,
which is projected with a linear layer to an embedding with size $d_p$.
The grasp pose $\mG$ input is a flattened rotation concatenated with the translation and embedded into ${\vz_{G} \in \R^{d_G}}$ using a linear layer with a Mish activation~\cite{DBLP:conf/bmvc/Misra20}
${f_{\text{G}}:\R^{9+3} \to \R^{d_G}}$.
The translation part of the grasp is normalized into the range $[-1, 1]$.
The denoising network $\vepsilon_{\vtheta}$ is a series of conditional residual blocks.
For each residual block, the condition variable $\vc$ is passed through a FiLM layer~\cite{DBLP:conf/aaai/PerezSVDC18} producing vectors $\va, \vb \in \R^{d_G}$ of an affine transformation.
The input to the residual block is the grasp embedding $z_G$, which is passed through a linear layer with a Mish activation before undergoing the affine transformation 
$\va f(\vz_{\mG}) + \vb$, 
and added to the original embedding.
These residual blocks are repeated $n_r=4$ times.
Finally, the last embedding is projected with a linear layer $f_{\text{out}}:\R^{d_G} \to \R^6$ to the output denoising vector $\vepsilon \in \R^6$.
The parameters $\vtheta$ include the point cloud encoder and the denoising networks and are learned jointly by optimizing the loss in \cref{eq:diffusion_loss_se3}.

Grasp poses are extremely sensitive to the translation component.
If the translation changes by $1$~cm, it can turn a previously successful grasp pose into failure.
This can happen when sampling from diffusion models because noise is added at every step.
As proposed by~\citet{DBLP:conf/iclr/AjayDGTJA23}, to control the level of stochasticity we use low-temperature sampling by multiplying the covariance with a scalar $\alpha \in [0, 1]$, and sample from
${
\IGSOthreeRthree ( \mG_{i-1}; \vmu_{\vtheta}(\mG_{i}, i, \vc), \alpha \mSigma_i)
}$.
We present an ablation study of this parameter in \cref{sec:ablation_low_temperature_sampling}.

\subsection{Object Collision Cost-Guided Grasp Diffusion}
\label{sec:object_collision_cost_guided_grasp_diffusion}

Along with a grasp generator network, previous work~\cite{DBLP:conf/iccv/MousavianEF19} proposed to use a grasp evaluator (a classifier) of \textit{good}/\textit{bad} grasps to generate grasps in a two-stage process.
Grasps are first sampled from the generator and then refined with the evaluator network using gradient ascent.
Grasps sampled from the generator were often in collision, and therefore, an evaluator network would be beneficial to improve those grasps.
However, training a classifier requires learning a second network, which needs further hyperparameter tunning.
Hence, we propose a more straightforward approach.
While evaluating the grasp generation, we noticed a negative correlation between the grasp success rate and the collision rate (fraction of grasps that are in collision with the object's partial point cloud).
Therefore, to encourage moving grasps to collision-free regions, we propose to approximate the gripper and the partial point cloud populated with collision spheres and optimize the collision cost function
\begin{align}
    C_{\text{coll-obj}}(\grasp) & = \int_{\mathcal{B}} C_{\text{obj}}(\vx_{m}(\grasp)) \dd m \approx \sum_{m=0}^{S-1} C_{\text{obj}}(\vx_{m}) \label{eq:cost_collision_avoidance}
\end{align}
where 
$
C_{\text{obj}}(\vx_m) = \sum_{j=0}^{N-1} \relu( -\| \vx_m - \vx_j \|  + r_m + r_j + \epsilon)
$,
$\vx_m \in \R^3$ is the position of the gripper's $m$th sphere center computed with forward kinematics at gripper configuration $\grasp$ and $r_m$ its radius, and $\vx_j \in \R^3$ is the position of the point cloud $j$th sphere center and $r_j$ its radius.
To prevent (and control) the gripper being \textit{too} close to the point cloud, we use a safety margin $\epsilon \geq 0$ and activate the collision cost only if the gripper is inside the safety margin.
One benefit of diffusion models is combining sampling from the prior with likelihood optimization by using classifier-guided diffusion~\cite{DBLP:conf/iros/Carvalho0BK023,Dhariwal2021diffusionbeatsgans,janner2022diffuser}.
We consider the previously trained diffusion model as a prior on \textit{good} grasps $p( \grasp_0 | \context)$, and propose to use classifier-guided diffusion to sample from the posterior distribution
$
p(\grasp_0 | \objective, \context) \propto p(\objective | \grasp_0, \context) p( \grasp_0 | \context),
$
where $\objective$ is an optimality variable, and the likelihood and cost are related by 
${p(\objective | \grasp_0, \context) \propto \exp \left( -C_{\text{coll-obj}} (\grasp) \right)}$~\cite{urain_2022_learning_implicit_priors,DBLP:conf/iros/Carvalho0BK023}.
If we had access to the full mesh of the object, we could, in principle, do gradient descent on the collision cost and get a grasp pose that is not in collision.
However, in practice, as we only have access to a partial point cloud, using this cost without the prior could lead to collisions with the unseen part of the object.
Hence, it is important to optimize the collision cost and stay close to the prior distribution.
This optimization is done in DDPM by, at every denoising step shifting the posterior mean with the cost function gradient using~\cite{Dhariwal2021diffusionbeatsgans}
\begin{align}
    & p_{\vtheta}(\mG_{i-1} | \mG_{i}, i, \context, \objective) = \nonumber \\
    & \qquad
    \IGSOthreeRthree ( \mG_{i-1}; \vmu_{\vtheta}(\mG_{i}, i, \context) \oplus \lambda \mSigma_i \vg, \mSigma_i)
    \label{eq:posterior_classifier_guidance} \\
    &
    \vg = -\nabla_{\mG} C_{\text{coll-obj}} (\mG) \rvert_{\mG = \vmu_{\vtheta}(\mG_{i}, i, \context)}, \label{eq:gradient_collision_cost}
\end{align}
where $\lambda$ is a step size. 
The collision cost function (\cref{eq:cost_collision_avoidance}) is computed with the spheres' positions, whose relative distances are fixed.
To keep their spatial relation,
we parametrize the gripper pose $\mG$ with $3$ prismatic and $3$ revolute joints, $\mG \coloneqq \vq = (\vp, \ve)$, where $\vp$ is the translation part, and $\ve$ is the rotation part represented in Euler angles. 
The posterior mean is combined with the (scaled) gradient using this parametrization and afterward converted back to $\mG$.
We note that $\mSigma_i$ decreases to $0$ when the denoising process evolves from $i \to 0$, so we divide the step size by the noise $\lambda \leftarrow \lambda \mSigma_{i}^{-1}$
to prevent the disappearing effect of the guidance function~\cite{DBLP:conf/iros/Carvalho0BK023}.
Because the gradient is only valid in a local neighborhood of $\vmu_{\vtheta}(\mG_{i}, i, \context)$, the step size is adapted such that the translation and rotation change by a maximum fixed $(\delta_p, \delta_r)$ amount, respectively, similar to~\cite{DBLP:conf/icra/ZhongRXCVCRP23}.
Finally, \citet{DBLP:conf/iclr/PearceRKBSGMTMH23} noticed that doing more denoising steps at time step $0$ increases the prediction accuracy, as this can be interpreted as moving samples towards higher-likelihood regions.
During gradient guidance, we employ both the step size adaptation and add more steps at denoising step $0$.

\section{Experiments}
\label{sec:experiments:gdn}
To assess the different components of GDN with simulation and real-world experiments, we pose the following research questions:
(1) Can we generate high-quality grasp distributions with GDN, and how does it compare to state-of-the-art methods?
(2) Can we accelerate grasp sampling without dropping the success rate?
(3) How does low-temperature sampling affect the results?
(4) How does the proposed collision avoidance cost guidance influence the success rate?
(5) Can we transfer the grasping generator to real-world scenarios?

\subsection{Dataset and Training Hyperparameters}
\label{sec:dataset}

To train our model, we use the ACRONYM dataset~\cite{DBLP:conf/icra/EppnerMF21}, consisting of $17.7$ million parallel grasps using the Franka Emika Panda gripper, from $8872$ objects and $262$ different categories.
We choose a subset of everyday objects consisting of $10$ categories
CAT10 = (Book,
Bottle,
Bowl,
Cap,
CellPhone,
Cup,
Hammer,
Mug,
Scissors,
Shampoo), 
totaling $567$ objects and $1.134$ million grasps.
Contrary to previous work~\cite{urain2022se3diffusion}, which trained separate models for each object category, training a single model is desirable in real-world scenarios where we might not have access to an object classifier.
We use the train/test splits from~\cite{DBLP:conf/icra/SundermeyerMTF21}.

The diffusion model uses $100$ diffusion steps and $\beta$ noise cosine schedule.
The joint context vector is $\vc \in \R^{256}$ and the grasp embedding $\vz_G \in \R^{256}$. 
The loss function is optimized with mini-batch gradient descent using the Adam optimizer~\cite{DBLP:journals/corr/KingmaB14} and a learning rate ${3 \times 10^{-4}}$.
For each training mini-batch, we render partial point clouds from a virtual camera of $32$ objects and use $32$ grasps per object (a total of $1024$ grasps per batch).
The point cloud is downsampled to $1024$ points and embedded into $\vz_p \in \R^{256}$.
See \cref{fig:gdn_method_overview} for a reference.
The models are trained for a maximum of $300$k epochs, or $48$ hours, with an NVIDIA RTX 3090 GPU.

\subsection{Baselines}
\label{sec:baselines:gdn}

We benchmark our method against the following baselines:
(1) GDN-Euclidean, which uses our proposed method with a continuous $6$d representation~\cite{DBLP:conf/cvpr/ZhouBLYL19} and performs diffusion in the Euclidean space, similar to diffusion for translations.
(2) A Conditional Variational Auto-Encoder (CVAE)~\cite{DBLP:conf/nips/SohnLY15}, sharing the same architecture of GDN but where the first $2$ blocks are used as an encoder to learn the mean and standard deviation of a Gaussian distribution in the latent space, and the last $2$ blocks are used to decode the latent space sample into a grasp pose.
This choice guarantees the number of parameters is similar to the diffusion-denoising network.
The latent space dimension is $4$.
The input to the CVAE model is the flattened rotation matrix concatenated with the translation, and the output is the reconstructed grasp in the form of the $6$d representation of rotation and the translation.
The loss function optimized is the \mbox{L$2$-norm} between the input and reconstructed grasps (with rotations has the $6$d representation) and a KL-divergence term between the latent space posterior distribution and the standard Gaussian, weighted with $10^{-2}$.
(3) $\SEthree$-DiffusionFields~\cite{urain2022se3diffusion}, which models the distribution with an energy-based model that is trained using denoising score-matching and establishes the state-of-the-art for grasp generation with score-based/diffusion models in $\SEthree$.

\subsection{Grasp Generation Quality in Simulation}
\label{sec:grasp_generation_simulation}

To evaluate the grasp generation quality, we report the success rate and earth mover's distance (EMD) between the generated and the ground truth data as in~\cite{urain2022se3diffusion}.
The success rate in simulation is computed with IsaacGym by placing the object floating, and the gripper is spawned at the generated grasp locations.
Then, the gripper is closed and lifted in the vertical direction, and success is achieved if the object remains between the gripper's fingers.
Each grasp generator is evaluated in $50$ objects from the test set with $5$ random rotations, producing different partial point clouds and $100$ grasps sampled from the models - a total of $250$ partial point clouds and $25$k grasps.

\begin{figure}[!t]
    \centering
    \includegraphics[width=0.95\columnwidth]{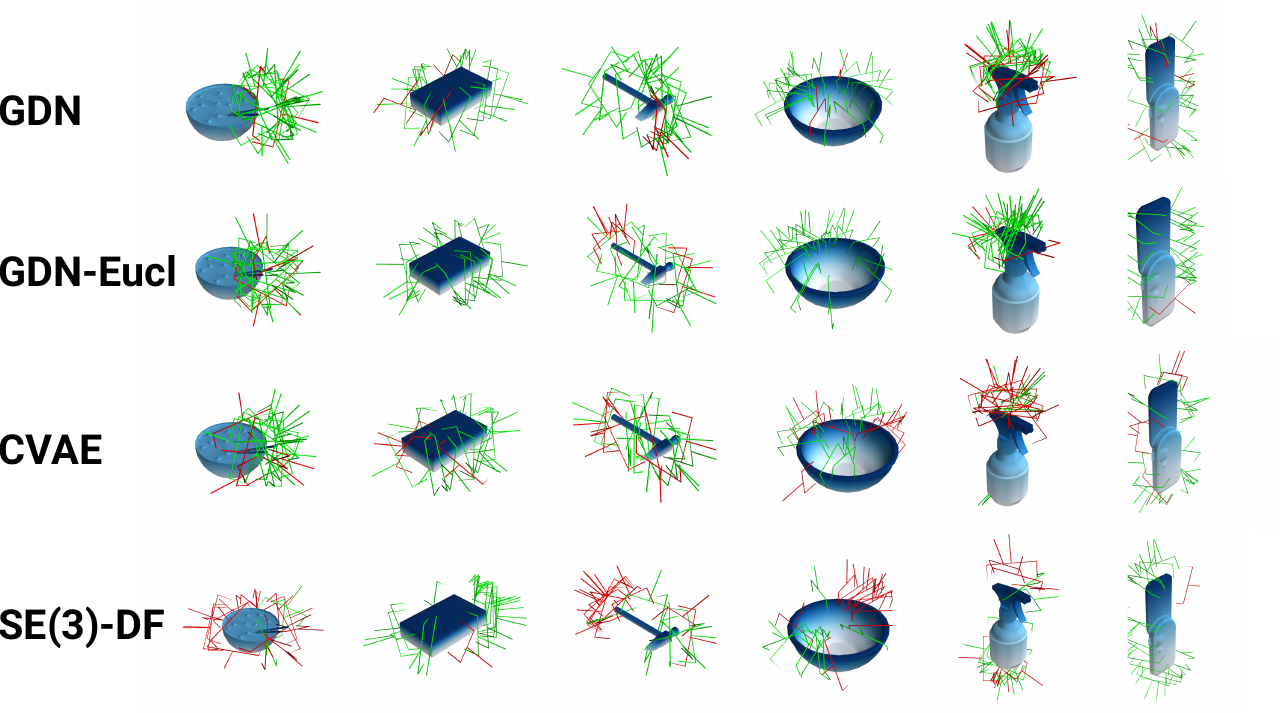}
    
    \caption{
        Examples of grasp samples generated with models trained on the CAT10 category.
        The $1$st and $5$th columns show more noticeable differences between the methods.
    }
    \label{fig:grasp_samples_generated}

    \vspace{-0.4cm}
\end{figure}

\begin{figure}[!t]
    \centering

    \includegraphics[width=0.44\columnwidth]{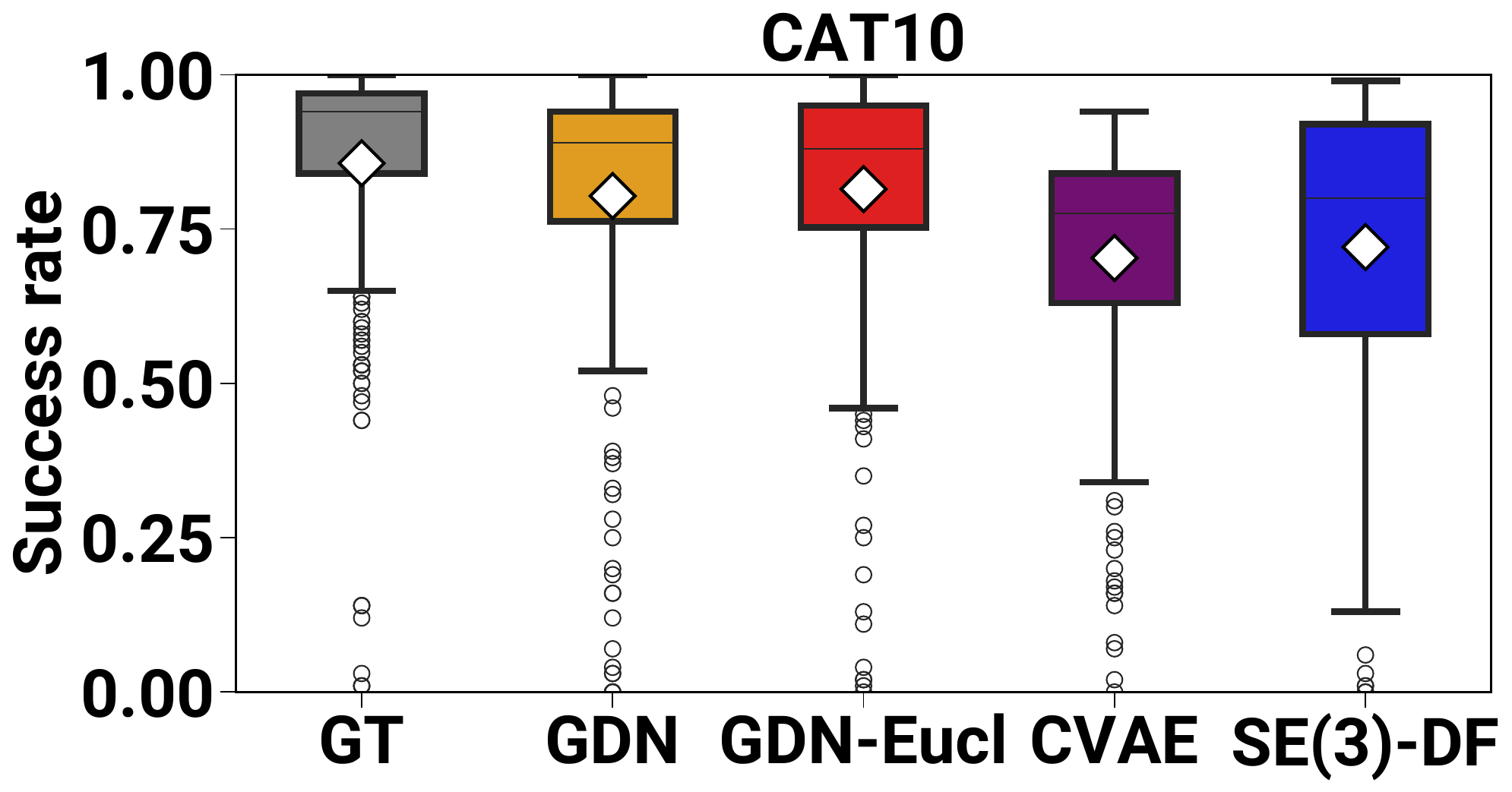}
    \hfill
    \includegraphics[width=0.44\columnwidth]{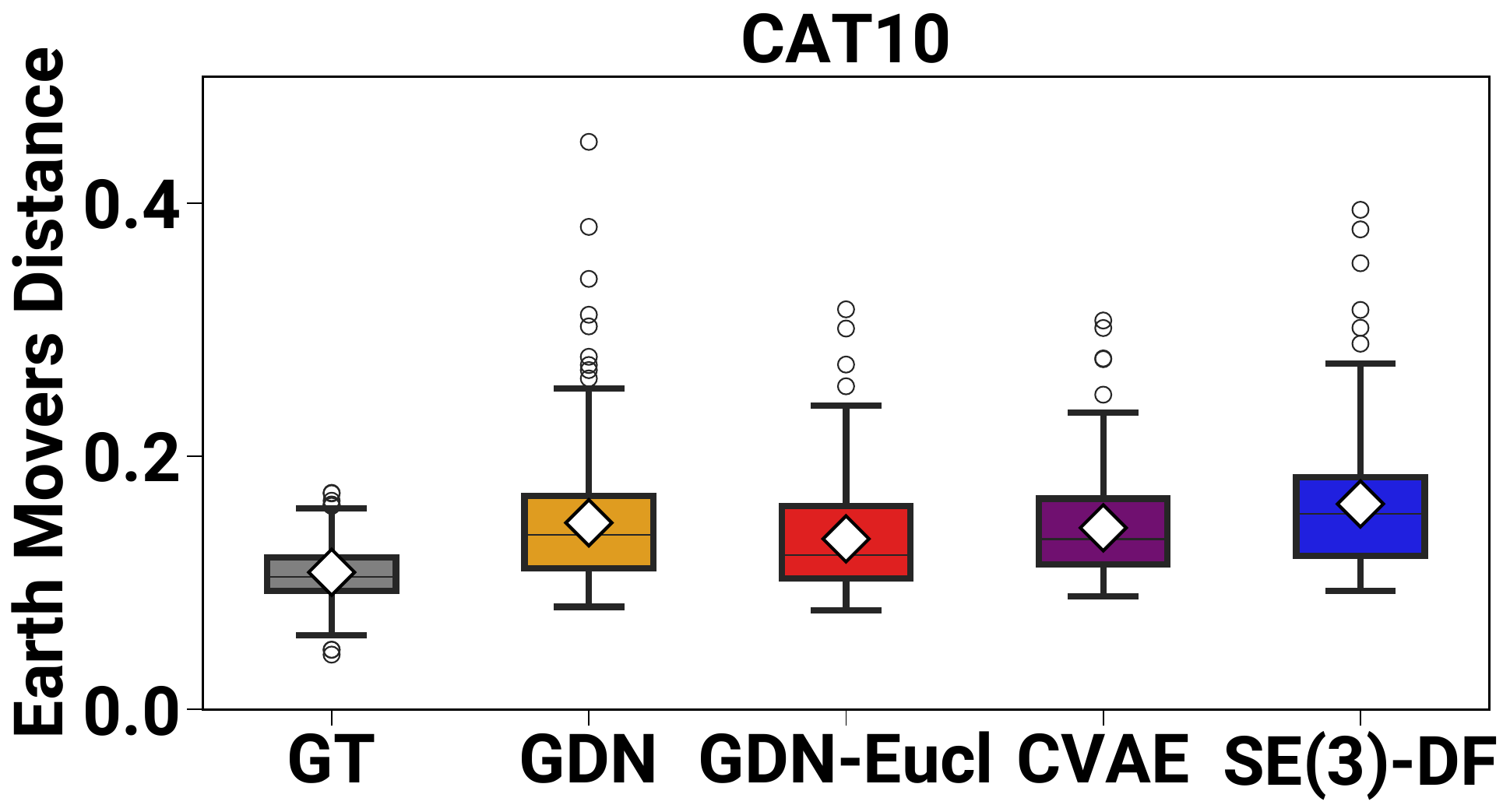}
    \hfill
    
    \caption[Grasp generation success rate and earth movers distance results for different categories and algorithms.]{
        Performance of different grasp generator models on CAT10 categories.
        The diamond in the boxplots shows the mean of the success rate or EMD.
        The results show that GDN can generate high-diversity grasps (low EMD) and precise ones (high success rate), either at the same level or better than the baselines.
    }
    \label{fig:simulation_results}
    
    \vspace{-0.5cm}
  
\end{figure}

\Cref{fig:grasp_samples_generated} shows examples of grasp samples for all the methods across different objects of the test set.
The evaluation results are visualized in \cref{fig:simulation_results}.
The ACRONYM dataset was generated with an NVIDIA Flex simulator, which is not publicly available.
Therefore, along with our method and baselines, we also report the ground truth statistics, which act as an upper-bound reference.
The success rate is not $1.0$ due to the imperfect transfer from Flex to IsaacGym.
The results show that GDN either matches the success rate or surpasses the baselines, showing also less variance in the results.
For instance, GDN's mean success rate is $80.35 \%$, compared to $70.31 \%$ and $72.08 \%$ for the CVAE and $\SEthree$-DiffusionFields baselines, respectively.
The lower EMD values also show that our method can generate diverse grasps and close to the ground truth distribution.
Interestingly, we observed that the CVAE baseline can also produce diverse grasps, but it showed a lower average success rate than our proposed approach. 
The results of GDN-Euclidean also show an interesting point.
Contrary to our expectations, this model performs as well as doing diffusion in the $\IGSOthree$ probability manifold.
For instance, the success rate in the CAT10 category is $81.47\%$.
The same phenomenon was noticed recently by~\citet{chisari2024learning} when using flow-matching for policy learning in the end-effector trajectory space.
We leave the theoretical analysis of this result for future work.
We also believe that our increased results in comparison to SE(3)-DiffusionFields might come from a simpler architecture and by modeling the score function directly with the denoising network output instead of computing it via the gradient of an Energy-Based Model.

From the generated samples in \cref{fig:grasp_samples_generated} we observe that samples produced with GDN are more diverse and precise than SE(3)-DiffusionFields.
In the first column a bowl is shown, whose successful grasps are located near the chopsticks, but SE(3)-DiffusionFields generates grasps around the bowl's rim.
GDN generates grasps only near the chopsticks, which are more precise, thereby increasing the success rate.
In the second column, SE(3)-DiffusionFields concentrates more grasps on one side of the book, while GDN produces grasps on all sides, hence showing more diversity.
The same is visible for the bowl in the fourth column.

\subsection{Accelerated Grasp Generation}
\label{sec:accelerated_grasp_ddim}

\begin{figure}[!t]
    \centering

    \includegraphics[width=0.48\columnwidth]{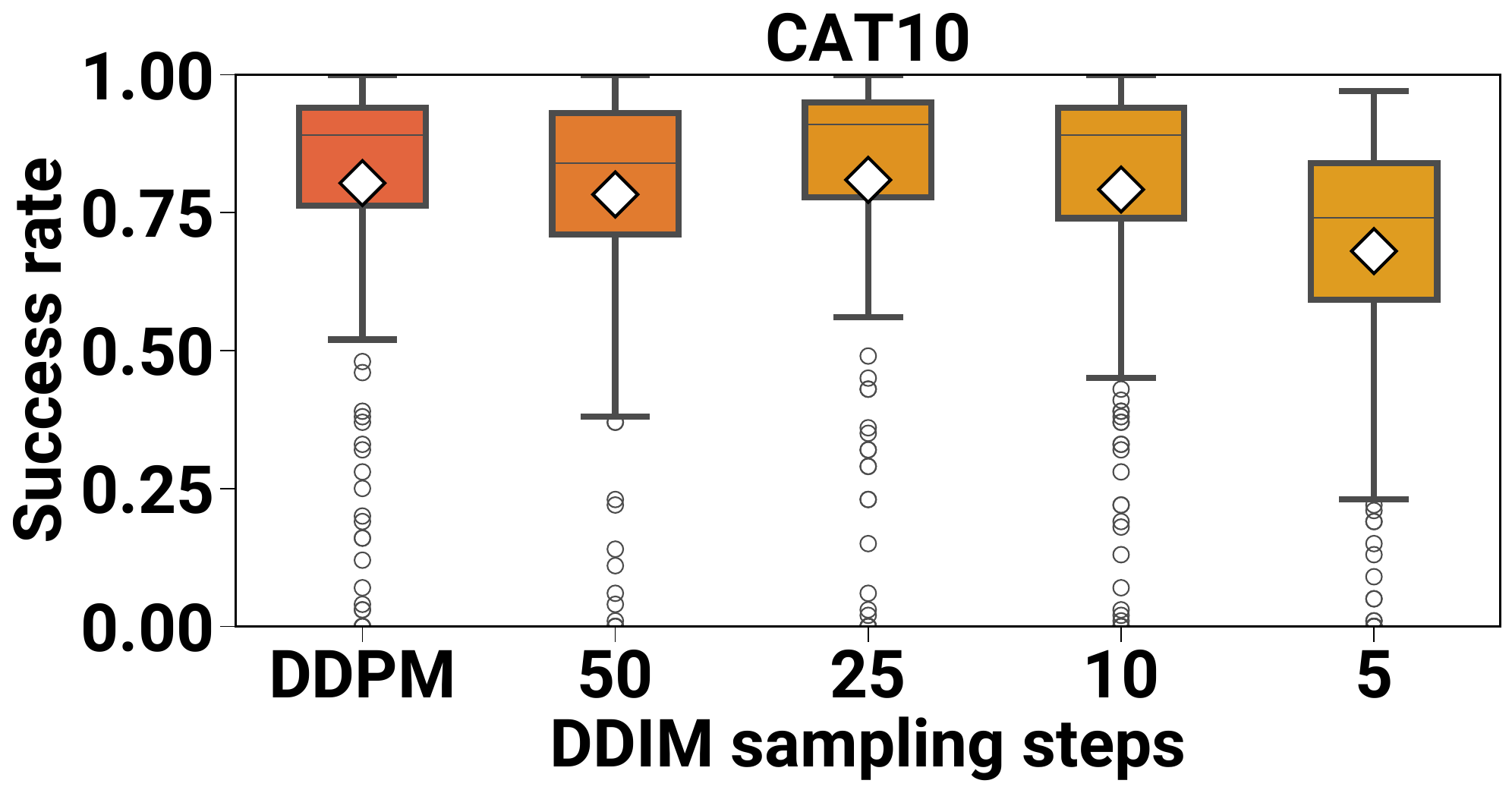}
    \hfill
    \includegraphics[width=0.48\columnwidth]{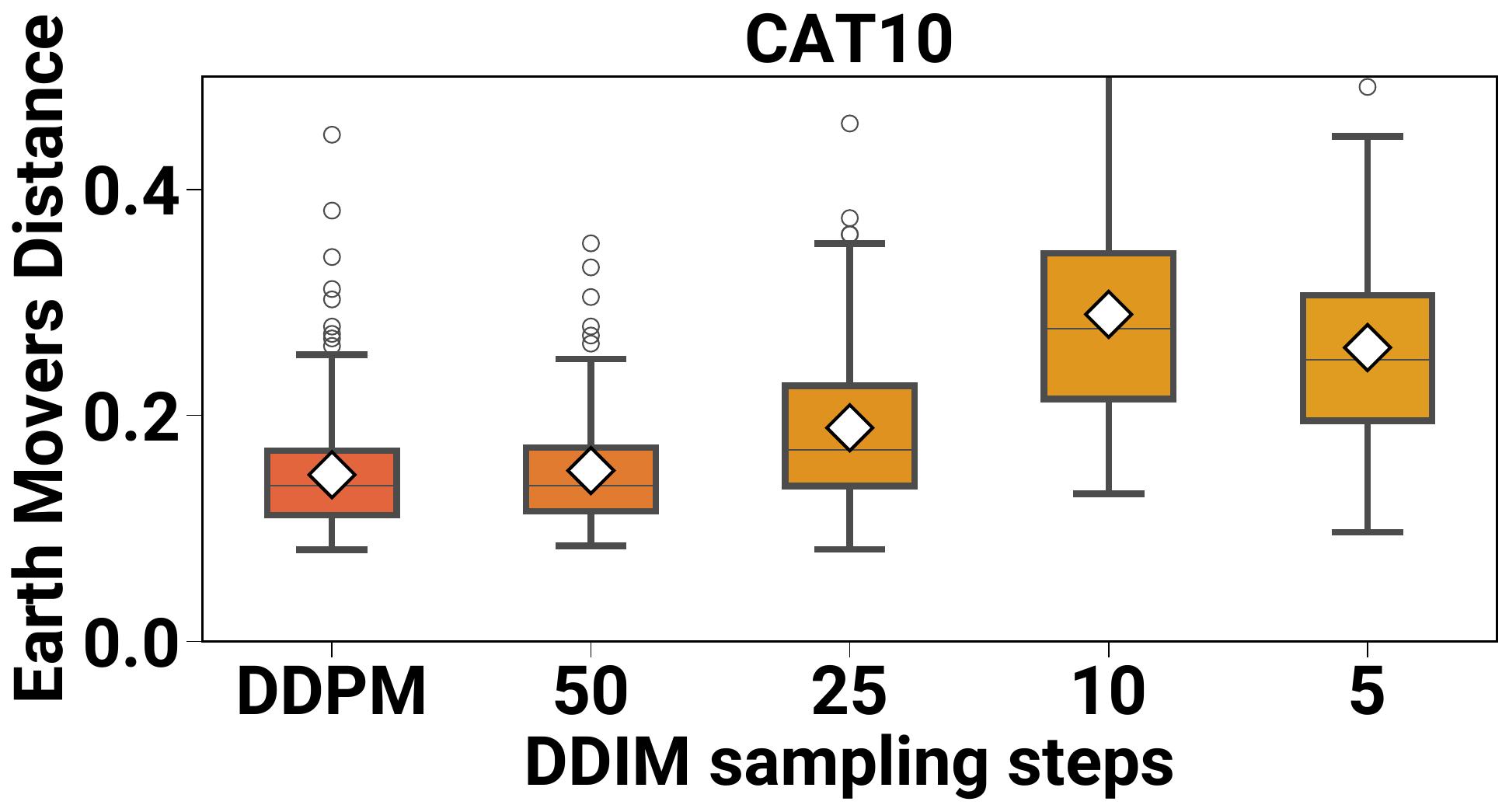}
    \hfill

    \caption[Grasp generation results using DDIM.]{
        GDN with fewer inference steps using DDIM.
        The diamond in the boxplots shows the mean of the success rate or EMD.
        The results show that several steps can be skipped during the denoising process without losing too much on grasp success rate but sacrificing grasp variability, as seen by the increase in EMD.
    }
    \label{fig:simulation_results_ddim_cat10}
    \vspace{-0.5cm}
\end{figure}

In our setup, sampling $100$ grasps with GDN using $100$ DDPM steps takes approximately $1.1$ seconds, while CVAE takes $0.02$ seconds since it only does one network pass, and $\SEthree$-DiffusionFields takes $1.6$ seconds.
Slow sampling speeds are undesirable for real-world applications.
Therefore, we propose to increase this speed by using DDIM and sampling with fewer time steps.
Instead of uniformly spaced indices, we propose to use a quadratic schedule, which focuses the denoising process on less noisy indices closer to $0$.
\Cref{fig:simulation_results_ddim_cat10} show that $10$ denoising steps provide a small drop in success rate at the expense of increasing the EMD, resulting in less diverse grasps.
The success rate with $10$ steps is $79.19\%$, which is still higher than the ones from CVAE and $\SEthree$-DiffusionFields ($70.31 \%$ and $72.08 \%$, respectively).
Moreover, the sampling time decreased from $1.2$ to only $0.2$ seconds, which can be further reduced by implementing faster $\IGSOthree$ samplers that we leave for future work.
In \cref{fig:grasp_samples_diffusion_ddim}, we render grasp poses sampled with DDPM and DDIM for the same objects, and is possible to observe that DDIM samples are more concentrated in certain regions of the objects.
For instance, in the third column, DDPM samples cover the bowl's rim, while DDIM samples are less spread.

\subsection{Low-Temperature Sampling}
\label{sec:ablation_low_temperature_sampling}

\Cref{tab:ablation_low_temperature_sampling} shows an ablation of the low-temperature sampling parameter (see \cref{sec:gdn_architecture_and_implementation_details}).
$\alpha = 1$ is the original DDPM version, and $\alpha = 0$ means no stochasticity during the denoising process.
The results show that the success rate increases with lower temperature sampling $\alpha \to 0$, but the EMD drops, indicating less stochasticity in the generated samples.
\begin{wraptable}{r}{0.3\textwidth}
    \vspace{-0.3cm}
    \footnotesize
    \caption[GDN results with different low-temperature sampling scales]{
        Metrics for GDN with DDPM and low-temperature sampling scales~$\alpha$.    
    }
    \centering
    \begin{tabular}{ccc}
    $\alpha$ & \textbf{Success rate} (\%) & \textbf{EMD} \\
    \hline
    $0.50$ & $82.38 \pm 23.38$ & $0.20 \pm 0.08$ \\
    $0.75$ & $80.35 \pm 22.46$ & $0.15 \pm 0.05$ \\
    $1.00$ & $75.48 \pm 22.52$ & $0.15 \pm 0.04$ \\
    \end{tabular}
    \label{tab:ablation_low_temperature_sampling}
    \vspace{-0.3cm}
\end{wraptable}
Given these results, we run GDN with DDPM using $\alpha = 0.75$ as a compromise between success rate and EMD.

\subsection{Collision-Cost Guidance}
\label{sec:experiments_collision_cost_guidance}

\begin{table}[!t]
    \footnotesize
    \caption[Results for collision-avoidance cost guided diffusion using GDN.]{
        Metrics for different sampling timesteps at diffusion index $0$ ($K$) and intermediate guide steps ($M$).
        GDN uses DDPM with $\alpha=0.75$.
    }
    \centering
    \begin{tabular}{ccccc}
    $K$ & $M$ & \textbf{Success rate} (\%) & \textbf{EMD} & \textbf{Collision rate} (\%) \\
    \hline
    \multicolumn{2}{c}{GDN} & $\bm{80.35} \pm 22.46$ & $ 0.147 \pm 0.050$ & $\bm{11.46}$ \\
    \hline
    $0$ & $1$ & $79.40 \pm 23.03$ & $0.151 \pm 0.052$ & $09.61$ \\
    $0$ & $2$ & $79.34 \pm 22.97$ & $0.151 \pm 0.052$ & $09.49$ \\
    $0$ & $3$ & $79.09 \pm 23.00$ & $0.151 \pm 0.052$ & $09.37$ \\
    \hline
    $1$ & $1$ & $81.01 \pm 22.78$ & $0.151 \pm 0.053$ & $08.76$ \\
    $1$ & $2$ & $81.02 \pm 22.77$ & $0.150 \pm 0.053$ & $08.76$ \\
    $1$ & $3$ & $80.83 \pm 22.80$ & $0.150 \pm 0.053$ & $08.71$ \\
    \hline
    $3$ & $1$ & $81.57 \pm 22.86$ & $0.150 \pm 0.053$ & $08.47$ \\
    $3$ & $2$ & $\bm{81.70} \pm 22.89$ & $0.150 \pm 0.053$ & $\bm{08.46}$ \\
    $3$ & $3$ & $81.34 \pm 23.00$ & $0.150 \pm 0.053$ & $08.42$ \\
    \end{tabular}
    
    \label{tab:abalation_cost_guidance}
    \vspace{-0.5cm}
\end{table}

In the simulation experiments, the average collision rate for GDN was $11.5\%$, while for the ground truth, it is $0.2\%$.
Importantly, the correlation between the success rate and the collision rate was $-0.45$, which indicates that lower collision rates might obtain higher success rates.
Several hyperparameters can be used during cost guidance.
In \cref{tab:abalation_cost_guidance}, we present the results by varying the number of steps at diffusion step $0$ ($K$) and the number of intermediate gradient steps ($M$) while modifying the step size such that the clipping values for maximum translation and rotation are $3$cm and $5^{\circ}$, respectively.
The results show that the proposed collision-avoidance cost guidance increases the success rate, particularly if more steps are done without adding noise to the grasp samples.
The best result was using $K=3$ steps at index $i=0$ and $M=2$ intermediate steps.
There is a maximum drop in collision rate of $3\%$ from $11.46\%$ to $8.46\%$, while the success rate increases $1.45\%$ from $80.35\%$ to $81.70\%$ (the ground truth is $85.6\%$), which is in line with the correlation value ($\approx -0.5$).
Even though there is a gain in using collision-avoidance guidance, the gain might be relatively small, suggesting the generator already models the grasp distribution quite well.

\subsection{Real-world Grasping Experiments}
\label{sec:real_world_experiments}

\begin{table}[!t]
        \footnotesize
        \centering

        \setlength{\tabcolsep}{4pt}
        \begin{tabular}{ l c c c c }
            \textbf{Objects} & GDN-DDPM & GDN-DDIM(10) & CVAE & \SEthree-DF \\
            \hline
            1.  Book & 4/5 & 5/5 & 5/5 & 4/5  \\
            2.  Shampoo & 3/5 & 3/5 & 0/5 & 5/5  \\
            3.  Bottle & 4/5 & 4/5 & 3/5 & 4/5  \\
            4.  Mug & 5/5 & 5/5 & 2/5 & 2/5 \\
            5.  Cup & 5/5 & 3/5 & 4/5 & 4/5 \\
            6.  Bowl & 5/5 & 5/5 & 5/5 & 2/5 \\
            7.  Cup & 5/5 & 5/5 & 5/5 & 3/5 \\
            8.  Bowl & 5/5 & 5/5 & 5/5 & 2/5 \\
            9.  Cell phone & 4/5 & 5/5 & 3/5 & 3/5 \\
            10. Cap & 3/5 & 5/5 & 1/5 & 3/5 \\
            \hline
            Success rate & $86.00\%  $ & $\bm{90.00}\%  $  &  $66.00\%  $  &  $64.00\%  $ \\
            \hline
        \end{tabular}
        \captionof{table}[Real-world grasping success rate results.]{
            Real-world successful grasps over the total number of executed grasps per object and algorithm for different objects.
        The object numbering corresponds to \cref{fig:real_world_objects}.
        }
        \label{tab:real-world-experiment}
    \vspace{-0.4cm}
\end{table}

\begin{wrapfigure}{r}{0.35\columnwidth}
    \vspace{-0.4cm}
    \centering
    \includegraphics[width=0.35\columnwidth]{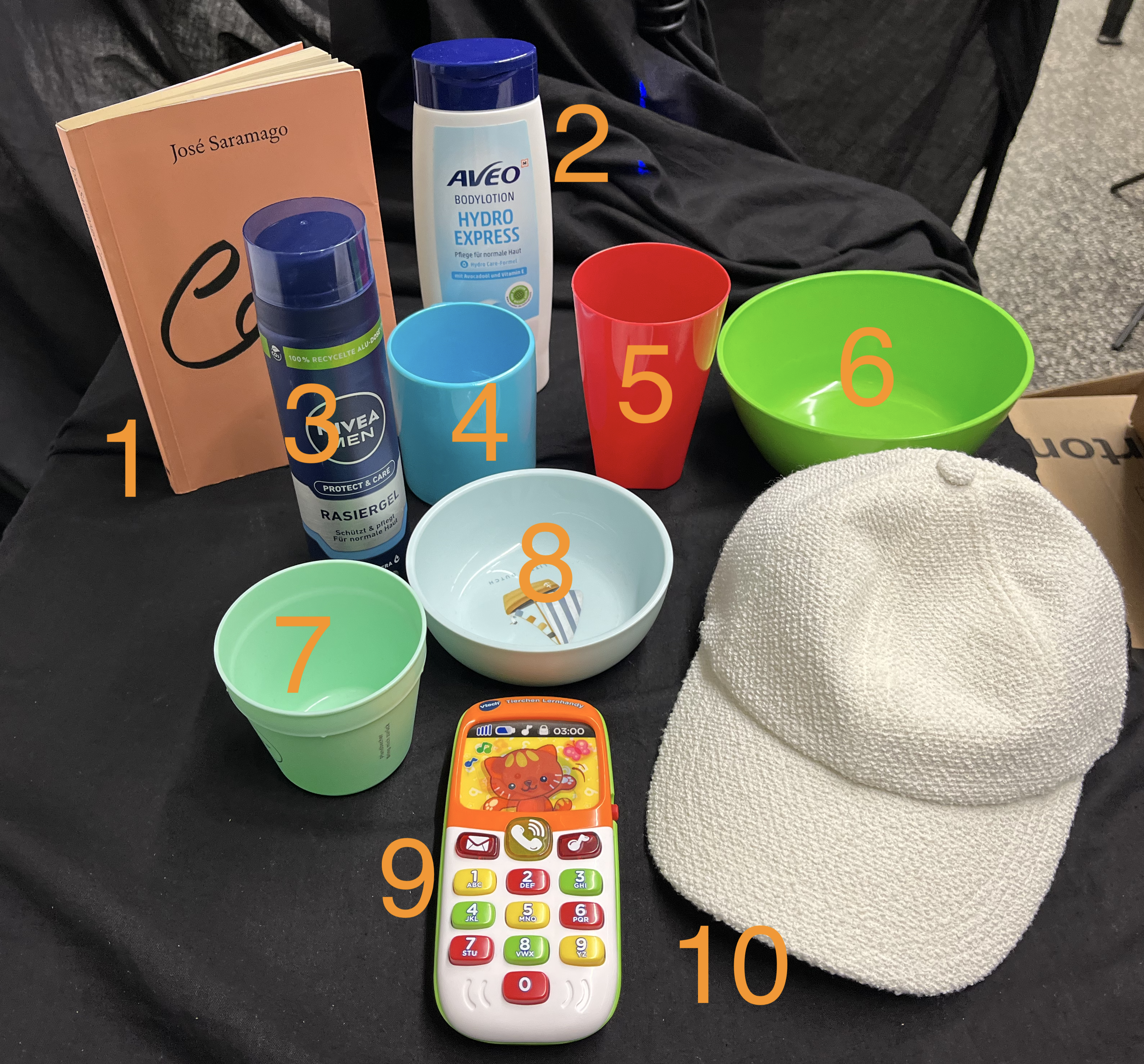}
    \caption[Objects used in the real-world grasping experiments.]{Objects used in the real-world experiments.}
    \label{fig:real_world_objects}
    \vspace{-0.5cm}
\end{wrapfigure}

In the last experiment, we evaluate how the grasp generative models' performance transfers from the simulation experiments to a real-world grasping scenario.
We test the following models:
GDN using DDPM, 
GDN using DDIM with $10$ sampling steps,
the CVAE baseline,
and the $\SEthree$-DiffusionFields baseline.
The real-world setup is shown in \cref{fig:grasping_real_world}.
We bought $10$ objects in a local store from the CAT10 categories, which are displayed in \cref{fig:real_world_objects}.
In this experiment, the Franka Emika Panda robot needs to grasp an object from a table and drop it in a box, using only a partial point cloud view from an external RGB-D camera (Azure Kinect).
We use MoveIt~\cite{DBLP:journals/corr/ColemanSCC14} for camera calibration and obtain the transformation between the camera and the robot base $\HBinA{r}{c}$. 
Given an RGB-D image, we use the state-of-the-art segmentation method Fast Segment Anything (FastSAM)~\cite{DBLP:journals/corr/abs-2306-12156} to obtain the segmented image of the object, from which we then retrieve the point cloud using the camera's intrinsic projection matrix, and remove outliers.
Alongside, we obtain the transformation from the camera frame to the point cloud center frame $\HBinA{c}{p}$.
We sample $100$ candidate grasp poses, which are 
represented in the point cloud center frame $\HBinA{p}{g}$,
and transform them to the robot's frame using
${\HBinA{r}{g} = \HBinA{r}{c} \, \HBinA{c}{p} \, \HBinA{p}{g}}$.
Afterward, we remove the grasps colliding with the partial point cloud and the table and select one randomly.
We use a path planner to move the robot to a pre-grasp and grasp joint goal configurations (computed with inverse kinematics) while avoiding collisions with the environment.
At the grasp pose, the gripper is closed, and the end-effector moves first $50$cm vertically and then to a disposal area, where the object is dropped into a box.
We consider a successful grasp if the gripper can lift the object and keep it in its fingers until reaching the box.

\begin{figure}[!t]
    \centering

    \includegraphics[width=0.32\columnwidth]{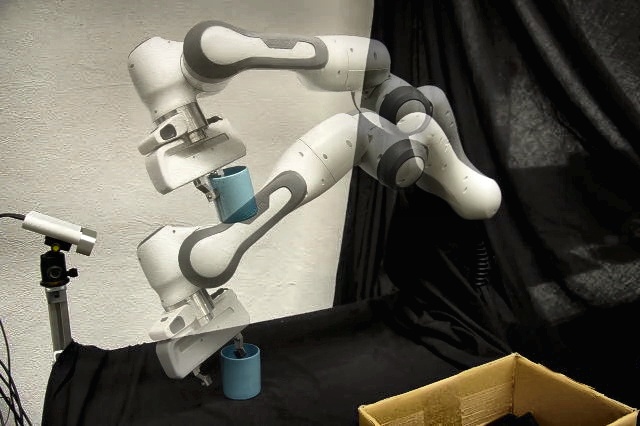}
    \hfill
    \includegraphics[width=0.32\columnwidth]{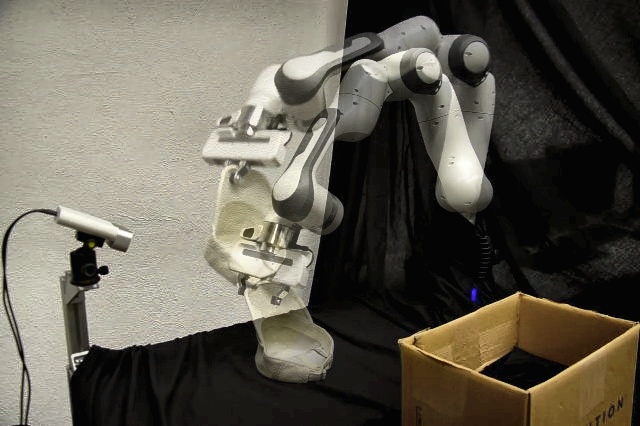}
    \hfill
    \includegraphics[width=0.32\columnwidth]{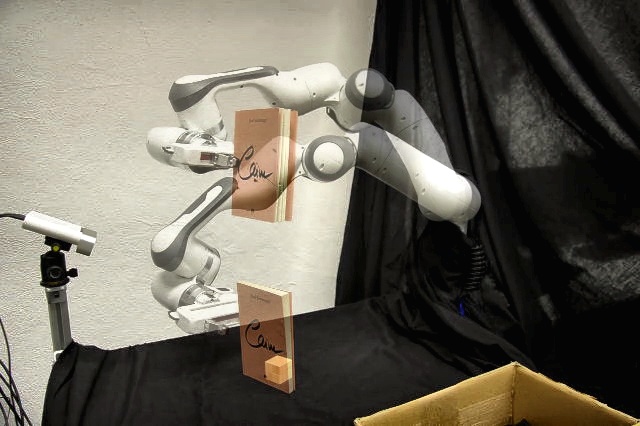}

  \caption[Successful execution of grasps in the real world scenario.]{
    The figures show overlays of the successful grasp execution sampled with GDN for different real-world objects.
  }
  \label{fig:real_world_rollouts}
  \vspace{-0.5cm}
\end{figure}

Each object is placed on the table in $5$ different poses, such that each model is evaluated on $50$ grasps.
In total, we perform $200$ real-world grasps.
\Cref{fig:real_world_rollouts} shows examples of successful grasps execution sampled with GDN for different objects, and the success rate is reported in \cref{tab:real-world-experiment}.
The results mirror those from the simulation experiments.
GDN shows consistently better results than the baselines across all objects.
In particular, the DDIM version of GDN, which allows for fast grasp generation times, provided the highest success rate.
One possible reason to justify this is that DDIM is a deterministic sampling method known to produce more precise samples,
which is crucial in the context of grasping since altering the grasp pose by $1$cm can determine whether the grasp is successful or not.
The baselines CVAE and $\SEthree$-DiffusionFields already showed in the simulation experiments lower success rates than GDN, which is also noticeable in the real-world task.

\section{Conclusion}
\label{sec:conclusion}

We proposed Grasp Diffusion Network (GDN), a new grasp generative model to sample grasp poses given partial point clouds of single objects.
Our method encodes the grasp distribution via a diffusion process in the $\SOthreeRthree$ probability distribution manifold of homogeneous transformations.
Along with the grasp generative model, we introduce a simple approach to increase the grasp success rate by using a collision-avoidance cost function during inference.
Using implicit diffusion sampling methods, we obtain faster sampling rates needed for real-world applications.
Our results in simulation show the benefits of the different components of our approach.
Furthermore, we perform experiments in a table-top grasping scenario using real-world objects, showing that GDN can transfer from simulation to the real world and obtain better success rates than the baselines, which are in line with the findings from the simulation experiments.

In future work, 
we will study the theoretical differences between diffusion in the $\SOthreeRthree$ or the Euclidean manifold in the context of grasp generation,
and expand our method to consider grasping objects in cluttered scenes.

\bibliographystyle{bibliography/IEEEtran}
\bibliography{bibliography/IEEEabrv,
bibliography/references
}

\end{document}